\documentclass{IEEEtran}
\usepackage[T1]{fontenc}
\usepackage{microtype}
\usepackage{graphicx}
\usepackage{subfigure}
\usepackage{booktabs} 

\usepackage{float}
\usepackage{algorithm}
\usepackage{algorithmic}

\usepackage{amsmath}
\usepackage{amssymb}
\usepackage{mathtools}
\usepackage{amsthm}

\usepackage{bbm}
\usepackage{amsfonts}       
\usepackage{amsthm} 

\theoremstyle{plain}
\newtheorem{theorem}{Theorem}[section]
\newtheorem{proposition}[theorem]{Proposition}

\theoremstyle{definition}

\theoremstyle{remark}
\newtheorem{remark}[theorem]{Remark}

\newcommand{\sign}[1]{\mbox{sign}(#1)}

\newcommand{\RR}{\ensuremath{\mathbb{R}}}

\newcommand{\calB}{\mathcal{B}}

\newcommand{\normone}[1]{{\|{#1}\|}_{1}}
\newcommand{\normtwo}[1]{{\|{#1}\|}_{2}}

\newcommand{\normOneInf}[1]{{\|{#1}\|}_{1,\infty}}
\newcommand{\normOneone}[1]{{\|{#1}\|}_{1,1}}
\newcommand{\normOnetwo}[1]{{\|{#1}\|}_{1,2}}
\newcommand{\norminf}[1]{{\|{#1}\|}_{\infty}}
\providecommand{\keywords}[1]
{
  \small	
  \textbf{\textit{Keywords---}} #1
}

\begin{document}

\title{A new Linear Time Bi-level $\ell_{1,\infty}$ projection ; 
           Application to the sparsification of auto-encoders neural networks }

\author{Michel Barlaud $^1 $, Guillaume Perez$^1 $ and Jean-Paul Marmorat $^2 $\\
$^1 $ I3S Laboratory, CNRS University of Côte d'Azur, France\\
$^2 $ CMA Laboratory, Ecole des Mines de Paris, France}

\vskip 0.3in
\maketitle

\\

\begin{abstract}
The $\ell_{1,\infty}$ norm is an efficient-structured projection, but the complexity of the best algorithm is, unfortunately, $\mathcal{O}\big(n m  \log(n m)\big)$ for a matrix in $\RR^{n\times m}$.\\
In this paper, we propose a new bi-level projection method, for which  we show that the time complexity for the $\ell_{1,\infty}$ norm is only $\mathcal{O}\big(n  m \big)$ for a matrix in $\RR^{n\times m}$.\\
Moreover, we provide a new $\ell_{1,\infty}$  identity with mathematical proof and experimental validation.\\ 
Experiments show that our bi-level $\ell_{1,\infty}$  projection is $\mathcal{O}\big(\log(n m)\big)$ times faster than the actual fastest algorithm. Our bi-level $\ell_{1,\infty}$  projection outperforms the sparsity of the usual $\ell_{1,\infty}$  projection while keeping the same  or slightly better accuracy in classification applications.
\end{abstract}

\section{Introduction}
Sparsity requirement appears in many machine learning applications,
such as the identification of biomarkers in biology \cite{he2010stable}.
It is well known that the impressive performance of neural networks is achieved at the cost of a high-processing complexity and large memory requirement \cite{AIgreen}.
Recently, advances in sparse recovery and deep learning have shown that training neural networks with 
sparse weights not only improves the processing time,
but most importantly improves the robustness and test accuracy of the learned models \cite{Alv2016,Han2015-2,Gom2019}\cite{dropout1,dropout2}. 
Regularizing techniques have been proposed to sparsify neural networks, such as the popular \textit{LASSO} method \cite{tRS,sparse}. The LASSO considers the  $\ell_1$ norm as Lagrangian regularization. 
 Group-LASSO originally proposed in \cite{Yua}, was used  in order to sparsify neural networks without loss of performance \cite{Hua2018,Yoo2017,Sca2017}. Unfortunately, the classical Group-LASSO algorithm is based on Block coordinate descent \cite{BCD,Fastlasso} and \textit{LASSO path} \cite{fht} which require high computational cost \cite{myCA} with convergence issue resulting in large power consumption. \\
An alternative approach is the optimization under constraint using projection \cite{Pesquet,BBCF}. Note that projecting onto the $\ell_1$ norm ball is of linear-time complexity \cite{condat,Perez19}. 
Unfortunately, these methods generally produce sparse weight matrices, but this sparsity is not structured  and thus is not computationally processing efficient. 
Thus, a structured sparsity is required (i.e. a sparsity able to set a whole set of columns to zero). 
The $\ell_{1,\infty}$ projection is 
of particular interest because it is able to set a whole set of columns to zero, 
instead of spreading zeros as done by the $\ell_1$ norm.
This makes it particularly interesting for reducing computational cost.
Many projection algorithms were proposed
\cite{quattoni2009efficient,bejar2021fastest}.
However, the complexity of these algorithms remains an issue. The worst-case time complexity of this algorithm is $\mathcal{O}\big(nm*\log(nm)\big)$ for a matrix in $\RR^{n\times m}$.
This complexity is an issue, and to the best of our knowledge, no current publication reports the use of the $\ell_{1,\infty}$ projection for sparsifying large neural networks. \\

The paper is organized as follows.
First, we provide the current state of the art of the $\ell_{1,\infty}$ ball projection. Then, we provide in section \ref{bi-level-1Infty} the  new bi-level $\ell_{1,\infty}$ projection. 
In section \ref{extend},  we apply our  bi-level framework to other  constraints providing sparsity, such as $\ell_{1,1}$ and $\ell_{1,2}$ constraints. In Section \ref{Exp}, we finally
compare different projection methods experimentally.
First, we provide an experimental analysis of the projection algorithms onto the  bi-level projection $\ell_{1,\infty}$ ball.
This section shows the benefit of the proposed method, especially for time processing and sparsity. 
Second, we apply our framework to the classification using a supervised autoencoder on two synthetic datasets and a biological dataset. 

\section{State of the art of the $\ell_{1,\infty}$ ball projection}
In this paper we use the following notations: lowercase Greek symbol for scalars, scalar i,j,c,m,n are indices of vectors and matrices, lowercase for vectors and capital for matrices. 

The $\ell_{1,\infty}$ ball projection has shown its efficiency to enforce structured  sparsity.\cite{quattoni2009efficient,chau2019efficient,chu2020semismooth,bejar2021fastest} and the classical approach is given as follows.\\
Let $Y \in \RR^{n \times m}$ be a real matrix of dimensions $ n \times m$, the elements of $Y$ are denoted by 
$Y_{i,j}$, $i=1,\ldots,n$, $j=1,\ldots,m$. 
The $\ell_{1,\infty}$ norm of $Y $ is
\begin{equation}
    \normOneInf{Y} := \sum_{j=1}^m \max_{i=1,\ldots,n} |Y_{i,j}|.
\end{equation}
Given a radius $\eta \geq 0$, the goal is to project $Y$ onto the $\ell_{1,\infty}$ norm ball of radius $\eta$, denoted by 
\begin{equation}
\calB^{1,\infty}_\eta:=\left\{X\in\RR^{n \times m}\ : \ \normOneInf{X}\leq \eta\right\}.
\end{equation}
The projection $P_{\calB^{1,\infty}_\eta}$ onto $\calB^{1,\infty}_\eta$ , also noted $P^{1,\infty}_\eta$ in the sequel, is given by:
\begin{equation}
P^{1,\infty}_\eta(Y)~=~P_{\calB^{1,\infty}_\eta}(Y) ~=~ {arg \min \limits_{X \in \calB^{1,\infty}_\eta}   \frac{1}{2} \|X-Y\|_\mathrm{F}^2} 
\end{equation}
where $\|\cdot\|_\mathrm{F}=\|\cdot\|_{2,2}$ is the Frobenius norm. \\
Let define the dual $\ell_{\infty,1}$ norm: 
\begin{equation}
\|Y\|_{\infty,1}:=\max_{j=1,\ldots,m}
\sum_{i=1}^n  |Y_{i,j}|.
\end{equation}
Given a matrix $Y\in\mathbb{R}^{n\times m}$ and a regularization parameter $\alpha >0$, the proximity operator of $\alpha \|\cdot\|_{\infty,1}$ is the mapping
\cite{moreau62}
\begin{equation}
\mathrm{prox}_{ \alpha \|\cdot\|_{\infty,1}}:Y\mapsto \arg \min \limits_{X\in\mathbb{R}^{n\times m}}   \frac{1}{2} \|X-Y\|_\mathrm{F}^2 + \alpha \|X\|_{\infty,1}.\label{eqprox}
\end{equation}

The proximity operator of the dual norm can be easily computed, then, using the Moreau identity \cite{moreau62,bau17,Combettes,con23} is an efficient method for computing the  projection onto the $\ell_{1,\infty}$ norm ball:
\begin{equation}
 P_{\calB^{{1,\infty}}_\alpha}(Y) =Y -\mathrm{prox}_{\alpha \|\cdot\|_{\infty,1}}(Y) 
\label{Moreau}
\end{equation}

A full description of the projection $P^{1,\infty}_\eta$, using Moreau identity and algorithm to compute it, can be found in \cite{Perez-2023,bejar2021fastest}.

\section{A new Bi-level  $\ell_{1,\infty}$ structured projection}

\label{bi-level-1Infty}
\subsection{A new bi-level projection}

In this paper, we propose the following alternative new bi-level method.
Let consider a matrix $Y $ with n rows and m columns.  Let  $y{_1}, \dots y{_m} $ the column vectors of matrix Y. 
Let ${v_\infty} = (\norminf{y_{1}},\dots,\norminf{y_{m}})$ the row vector composed of the infinity norms of the columns of matrix $Y$.
The bi-level $\ell_{1,\infty}$ projection optimization problem is defined by:
\begin{equation}
\begin{aligned}
BP^{1,\infty}_\eta(Y)=\{X|\forall j, x_j =\arg \min\limits_{x \in \calB_{\hat{u}_j}^{\infty}} \normtwo{x-y_j}\\
\text{such that}\quad 
\hat{u} \in \arg \min\limits_{u \in \calB_\eta^{1}} \normtwo{u-v_{\infty}}\}
\end{aligned}
\end{equation}
This problem is composed of two problems. 
The first one, the inner one, is:
\begin{equation}
\hat{u} \in \arg \min\limits_{u \in \calB_\eta^{1}} \normtwo{u-v_{\infty}}
\end{equation}
Once the columns of the matrix have been aggregated to a vector $v_{\infty} $ using the $\infty$ norm,
the problem becomes a usual $\ell_{1}$ ball projection problem.
The row vector $\hat{u}$ is given by the following projection of row vector v:
\begin{equation}
    \hat{u} \gets P^{1}_\eta((\norminf{y_{1}},\dots,\norminf{y_{m}}))
\end{equation}

\begin{remark}\label{p1_proj}
  As a contracting property of the $P^{1}_\eta$ projection, we have:\\
  \begin{equation}
 \norminf{y_{j}} \geq \hat{u}_j  \geq 0  \quad \forall j\in 1,\dots,m
  \end{equation}
  These bounds on the $u_j$ hold whatever the norm of the columns $y_j$.
\end{remark}
Then, the second part of the bi-level optimization problem, once the row vector $\hat{u}$ is known, is given by:
\begin{equation}
x_j = \arg \min\limits_{x \in \calB_{\hat{u}_j}^{\infty}} \normtwo{x-y_j}
\end{equation}
For each column $y_j$ of the original matrix, we compute an estimated column   $x_j$.
Each column $x_j$.is optimally computed using the projection on the $\ell_{\infty}$ ball of radius $\hat{u}_j$:
\begin{equation}
x_j \gets P^{\infty}_{\hat{u}_j}(y_j) \quad \forall j\in 1,\dots,m
\label{Projection}
\end{equation}
which can be written as  
\begin{equation}
X_{i,j} = \sign{Y_{i,j}} \min(|Y_{i,j}|,\hat{u}_j).
\label{norminfty}
\end{equation}

\begin{remark}
We say that $Y \rightarrow BP^{1,\infty}_\eta$ is a {\em clipping} operator, and $\hat{u}$ is its clipping threshold.
\begin{equation} 
\label{linf_error} 
\begin{aligned}    
Y_{i,j} - X_{i,j} &= \sign{Y_{i,j}} ( |Y_{i,j}| - \min(|Y_{i,j}|,\hat{u}_j)\\
|Y_{i,j} - X_{i,j}| &= | |Y_{i,j}| - \min(|Y_{i,j}|,\hat{u}_j)| 
\end{aligned}
\end{equation}
and then with remark~\ref{p1_proj}
\begin{equation}
\forall j \quad \max_i |Y_{i,j} - X_{i,j}| ~=~ \max_i |Y_{i,j}| - \hat{u}_j  ~=~  \norminf{y_{j}} - \hat{u}_j 
\end{equation}
Note by \ref{Projection} that  $ \hat{u}_j = \norminf{x_j}  $, so
\begin{equation}\label{basicInf}
\forall j \quad \norminf{y_j~-~x_j} ~=~\norminf{y_j} ~-~ \norminf{x_j}
\end{equation}

\end{remark}

Algorithm~\ref{algo:linfbiproj} is a possible implementation of $BP$.
It is important to remark that usual bi-level optimization requires many iterations \cite{bilevel2,bilevel3}
while our model reaches the optimum in one iteration.

\begin{algorithm}
   \caption{Bi-level $\ell_{1,\infty}$ projection ($BP^{1,\infty}_\eta(Y)$).}\label{algo:linfbiproj}
\begin{algorithmic}  
\STATE \textbf{Input:} $Y,\eta$
\STATE{$u \gets P^{1}_\eta((\norminf{y_{1}},\dots,\norminf{y_{j}},\dots,\norminf{y_{m}}))$}  
\FOR{$j \in [1,\dots,m]$}
  \STATE $x_j \gets P^{\infty}_{u_j}(y_j)$ 
\ENDFOR
\STATE \textbf{Output:} $X$
\end{algorithmic}
\end{algorithm}

\subsection{The $\ell_{1,\infty}$  identity }
In the case of $\ell_{1,\infty}$ projection, we needed Moreau's identity to develop the projection algorithm from the "Prox". 
In the case of our new bilevel $\ell_{1,\infty}$ projection, we have a direct linear-complexity algorithm that does not require Moreau's identity. 
The aim of this section is to show the respective properties of these two projections. We study a norm of the projected regularized solution versus a norm of the corresponding residual \cite{Lcurves}.
Recall the classical  triangle inequality, which is a consequence of the Cauchy–Schwartz inequality, 
\begin{equation}
 \normtwo{ Y - BP^{1,\infty}_\eta(Y)}+\normtwo{ BP^{1,\infty}_\eta(Y)} \geq \normtwo{Y}\\
\end{equation}

  However, we propose the following norm identity for the bilevel  $\ell_{1,\infty}$  projection.
\begin{proposition}
    
In the case of the $\ell_{1,\infty}$ norm, 
bilevel projected data and residual  are linked by the following relation:
\end{proposition}
\begin{equation} 
 \normOneInf{ Y - BP^{1,\infty}_\eta(Y)}+\normOneInf{ BP^{1,\infty}_\eta(Y)} =\normOneInf{Y}\\
 \label{Identity} 
\end{equation}

The proof of equation~\ref{Identity} is readily obtained by summation in $j$ of equation \ref{basicInf}.

\begin{remark}
Careful examination of the $P^{1,\infty}_\eta$ projector algorithm \cite{Perez-2023,bejar2021fastest,quattoni2009efficient} shows that $P^{1,\infty}_\eta$  is  also obtained by a clipping operator, for a different threshold $u$ (See Line 15 of algorithm 1 in \cite{Perez-2023}) and thus projection verifies  \ref{basicInf}.
 \end{remark}

 \begin{proposition}
The usual $P^{1,\infty}_\eta$ projection has the following property:
 \begin{equation}
 \normOneInf{ Y - P^{1,\infty}_\eta(Y)}+\normOneInf{ P^{1,\infty}_\eta(Y)} =\normOneInf{Y}\\
 \label{IdentityP} 
 \end{equation}
\end{proposition}

 The proof of Eq~\ref{IdentityP} follows the same way as for  Eq~\ref{Identity}.  In fact, identities such as \ref{Identity} and \ref{IdentityP} hold for infinitely many clipping operators. A vector $u$ is a feasible clipping  threshold if it satisfies bounds of remark~\ref{p1_proj} and sum to $\eta$.

\begin{remark}

Among all clipping operators, 
$P^{1,\infty}_\eta$ and $BP^{1,\infty}_\eta$ have the best properties for our purpose.  $BP$ has the best structured sparsification effect while  $P$ has the best $L_2$ error, However $L_2$ error is not more relevant for our purpose than any other norm (for example for the norm, $\ell_1,\infty$ BP and P provide the same error. 
\end{remark}

\subsection{Convergence and Computational complexity}
The best computational complexity of the projection of a matrix in $\RR^{nm}$ onto the $\ell_{1,\infty}$ ball is usually $O(nm\log(nm))$ \cite{quattoni2009efficient,bejar2021fastest}.\\
Our bilevel algorithm is split  in 2 successive projections. These projections give us a direct solution without iteration, so our algorithm converges in one loop.\\
The first projection is a $\ell_1$ projection applied to the m-dimensional vector of column norms; its complexity is therefore O(m) \cite{condat,Perez19}.
The second part is a loop (on the number of columns) of the $\ell_\infty$ projection, which is implemented with a simple clipping, so its complexity is $O(nm)$. 
Therefore, the  computational complexity of the bi-level projection here is $O(nm)$.

\section{Extension to other sparse structured projections  }

Let recall that there is a close connection both between the proximal operator \cite{Moreau}
of a norm and its dual norm, as well as between proximal operators of
norms and projection operators onto unit norm balls (pages 187-188, section 6.5 of \cite{Boyd}).

In this section, we extend our bilevel method  to the $\ell_{1,1}$ and $\ell_{1,2}$ balls, yielding structured sparsity.

\label{extend}
\subsection{Bilevel $\ell_{1,1}$ projection}

Let ${v_1} = (\normone{y_{1}},\dots,\normone{y_{m}})$ the row vector composed of the $\ell_1$ norm of the columns of the matrix $Y$.
We propose to define the $\ell_{1,1}$ bi-level optimization problem:

\begin{equation}
\begin{aligned}
BP^{1,1}_\eta(Y)=\{X|\forall j, x_j = \arg \min\limits_{x \in \calB_{\hat{u}_j}^{1}} \normtwo{x-y_j}\\
\text{such that}\quad 
\hat{u} \in \arg \min\limits_{u \in \calB_\eta^{1}} \normtwo{u-v_{1}}\}
\end{aligned}
\end{equation}

A possible implementation of the bi-level $\ell_{1,1}$ is given in Algorithm~\ref{algo:li1iproj}.

\begin{algorithm}[ht]
   \caption{Bi-level $\ell_{1,1}$ projection. ($BP^{1,1}_\eta(Y)$)}\label{algo:li1iproj}
\begin{algorithmic}  
\STATE \textbf{Input:} $Y,\eta$
\STATE{$u \gets P^{1}_\eta((\normone{y_{1}},\dots,\normone{y_{m}}))$} 
\FOR{$j \in [1,\dots,m]$}
  \STATE $x_j \gets P_{u_j}^{1}(y_j)$
\ENDFOR
\STATE \textbf{Output:} $X$
\end{algorithmic}
\end{algorithm}

Consider the $P_{u_j}^{1}$ projection on the $\ell_1$ ball of radius $u_j$in $\mathbb{ R}^n:$

The projection $x_j ~=~ P_{u_j}^1(y_j)$ is obtained by elementwise soft thresholding   (proposition 2.2 in \cite{condat} or section 6.5.2 in \cite{Boyd}), that is,
there exists some  positive unique $\lambda_j$ verifying a critical equation  such that:

\begin{equation*}
x_j = \max(y_j-\lambda_j, 0) - \max(-y_j - \lambda_j, 0)  
\end{equation*}
and  
\begin{equation*}
|x_j|_1 = u_j 
\end{equation*}
so
\begin{equation}
Y_{i,j} ~-~ X_{i,j} ~=~ \sign{Y_{i,j}} \left| ~ |Y_{i,j}| ~-~ \max( |Y_{i,j}| - \lambda_j, 0)) \right|
\end{equation}
and
\begin{equation}
\begin{aligned}
| Y_{i,j} ~-~ X_{i,j} |  &=   | ~ |Y_{i,j}| - \max(|Y_{i,j}|-\lambda_j,0)~| \\
 &=   |Y_{i,j}| -  \max(|Y_{i,j}|-\lambda_j,0)  \\
  &=   |Y_{i,j}| - |X_{i,j}|
\end{aligned}
\end{equation}

which can be written, by summing on $i$:
\begin{equation}\label{basicOne}
\forall j \quad \normone{y_j~-~x_j} ~=~\normone{y_j} ~-~ \normone{x_j}
\end{equation}

By direct summation on $j$ of equation \ref{basicOne}, we have:
\begin{proposition}
The  bilevel $\ell_{1,1}$ projection satisfies   the following identity.

\begin{equation} 
 \normOneone{Y - BP^{1,1}_\eta(Y)}+\normOneone{ BP^{1,1}_\eta(Y)} =\normOneone{Y}\\
 \label{Identity-L11} 
\end{equation}
\end{proposition}

\subsection{ Bilevel $\ell_{1,2}$ projection.}

Let ${v_2} = (\normtwo{y_{1}},\dots,\normtwo{y_{m}})$ the row vector composed of the $\ell_2$ norm of the columns of the matrix $Y$.
The $\ell_{1,2}$ bi-level optimization problem be:
\begin{equation}
\begin{aligned}
BP^{1,2}_\eta(Y)=\{X|\forall i, x_i = \arg \min\limits_{x_i \in \calB_{\hat{u}_i}^{2}} \normtwo{x_i-y_i}\\
\text{such that}\quad 
\hat{u} \in \arg \min\limits_{u \in \calB_\eta^{1}} \normtwo{u-v_{2}}\}
\end{aligned}
\end{equation}

Similarly, the bi-level projection algorithms for $\ell_{1,2}$ is given by algorithm~\ref{algo:li12proj}.
\begin{algorithm}[ht]
   \caption{Bi-level $\ell_{1,2}$ projection. ($BP^{1,2}_\eta(Y)$)}\label{algo:li12proj}
\begin{algorithmic}  
\STATE \textbf{Input:} $Y,\eta$
\STATE{$u \gets P^{1}_\eta((\normtwo{y_{1}},\dots,\normtwo{y_{m}}))$} 
\FOR{$j \in [1,\dots,m]$}
  \STATE $x_j \gets P^{2}_{u_j}(y_j) $
\ENDFOR
\STATE \textbf{Output:} $X$
\end{algorithmic}
\end{algorithm}


Let, $x_j ~=~ P_{u_j}^2(y_j)$   then, $x_j = \frac{u_j}{\|y_j\|_2} y_j $   (section 6.5.1 in \cite{Boyd})\\
and so $y_j ~-~ x_j ~=~ (1 ~-~ \frac{u_j}{\|y_j\|_2}) y_j$
then 
\begin{equation}\label{basicTwo}
\forall j \quad  \|y_j~-~x_j\|_2 ~=~ \|y_j\|_2 ~-~ u_j~=~\normtwo{y_j} ~-~ \normtwo{x_j}
\end{equation}
which by a direct summation on $j$ leads to

\begin{proposition}
The  bilevel $\ell_{1,2}$ projection satisfies   the following identity.
\begin{equation} 
 \normOnetwo{Y - BP^{1,2}_\eta(Y)}+\normOnetwo{ BP^{1,2}_\eta(Y)} =\normOnetwo{Y}\\
 \label{Identity-L12} 
\end{equation}
\end{proposition}



\section{Experimental results} 
\label{Exp}
\subsection{Benchmark times using PyTorch C++ extension using a MacBook Laptop with an i9 processor; Comparison with the best actual projection method }
The experiments were run on a laptop with  a I9 processor having 32 GB of memory.
The state of the art on such is pretty large, starting with \cite{quattoni2009efficient} who proposed the first algorithm, the Newton-based
root-finding method and column elimination method \cite{chau2019efficient,bejar2021fastest}, and the recent paper of \textit{Chu et al.} \cite{chu2020semismooth} which outperforms all the other state-of-the-art methods.
We compared our bi-level method against all the existing projection algorithms,
and our algorithm is faster in all the scenarios.
The algorithm from \cite{quattoni2009efficient} is on average 30 times slower than our algorithm,
note that this factor growth logarithmically with data size.
As shown in \cite{chu2020semismooth}, the best actual algorithm is proposed by \textit{Chu et al.}
which uses a semi-smooth Newton algorithm for the projection.
That is why we focus our presentation on comparing against this particular method.
We use the C++ implementation provided by the authors and 
the PyTorch C++ implementation of our bi-level $\ell_{1,\infty}$ method is based on fast
$\ell_{1}$ projection algorithms of \cite{condat,Perez19} which are of linear complexity.
The code of all the compared algorithms is available online\footnote{https://github.com/memo-p/projection}.\\


\begin{figure}[t]!
    \centering
    \includegraphics[width=0.49\textwidth,height=4.2cm]{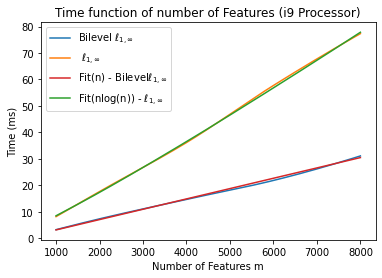}
    \includegraphics[width=0.49\textwidth,height=4.2cm]{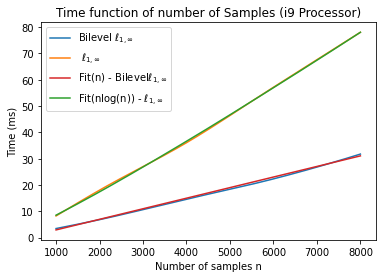}
    \caption{Processing time using C++ as a function of the number of features $n=1000$  samples (top) and Samples $m=1000$ features (bottom): bi-level projection method versus \textit{Chu et al.} method.}
    \label{Chu versus bilevel}
\end{figure}

Figure~\ref{Chu versus bilevel} shows the running time as a function of the matrix size. 
Here the radius has been fixed to $\eta=1$.
The fitting of a linear curve (red) on the data (blue) shows that the running time of our bilevel $\ell_{1,\infty}$ projection is linear with the number of features and the number of samples. 
The fitting of a $nlogn $ curve (green) on the data (orange) shows that the running time of the usual $\ell_{1,\infty}$ projection grows as $nlogn$ with the number of features and the number of samples. Moreover, the slope of the usual projection algorithm is greater by a factor 2.5 on both graphs than the slope of the bilevel algorithm.

\begin{figure}[t]
    \centering
    \includegraphics[width=0.49\textwidth,height=4.5cm]{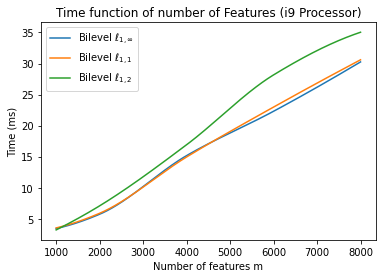}
    \includegraphics[width=0.49\textwidth,height=4.5cm]{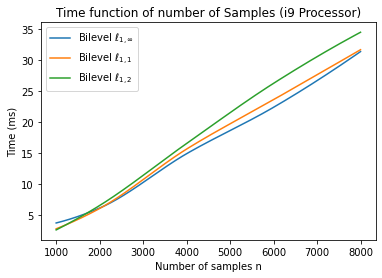}
    \caption{Processing time using C++ as a function of the number of features (Top), and samples (bottom)}
    \label{Bilevels}    

\end{figure}

Figure \ref{Bilevels} shows that all bilevel algorithms have the same slopes for time versus feature or sample number.

Note that PyTorch c++ extension is 20 times faster than the standard PyTorch implementation.

\subsection{Benchmark of Identity Proposition  }

We generate two artificial biological datasets to benchmark our bi level projection using  the $make\_classification$ utility from \textit{scikit-learn}. 
We generate $n=1,000$ samples  with a  number of $m = 1,000$ features. The first one with 64 (data-64) informative features and the second (data-16) with 16 informative features\\
We provide the experimental proof of the proposition and the sparsity score in $ \%$: number of columns or features set to zero.

\begin{figure}[t]
    \centering
    \includegraphics[width=0.49\textwidth,height=4.2cm]{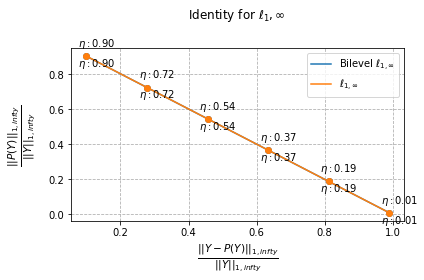} 
    \includegraphics[width=0.49\textwidth,height=4.2cm]{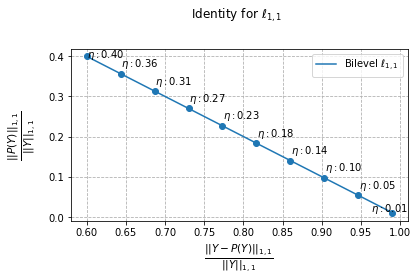}     
    \includegraphics[width=0.49\textwidth,height=4.2cm]{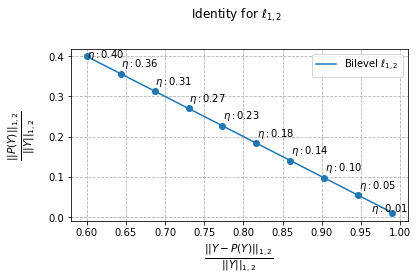}  
    \caption{Identity norm  comparison Top: the Bilevel $\ell_{1,\infty}$ versus classical, Middle: Bilevel $\ell_{1,1}$, bottom: Bilevel $\ell_{1,2}$ projection. }
    \label{NormL1infty}
\end{figure}
Fig \ref{NormL1infty} shows that the two curves  (Bilevel and usual $\ell_{1,\infty}$ projections)  and the $\eta$ parameter are perfectly coincident, and perfectly linear as expected by the identity equation (\ref{Identity}).

\begin{figure}[t]
    \centering
    \includegraphics[width=0.49\textwidth,height=4.2cm]{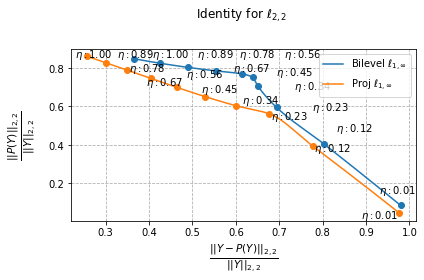} 
    \caption{ Bilevel $\ell_{1,\infty}$ projection and usual $\ell_{1,\infty}$ projection with $\ell_{2,2}$ norm.}
    \label{NormL1infty_L2norm}
\end{figure}
\begin{remark}
The identity equations hold only if the norm is similar to the projection.
Figure \ref{NormL1infty_L2norm} shows that the identity equation is not true when using Bilevel and usual $\ell_{1,\infty}$ projections and the classical $\ell_{2,2}$ norm. Usual projection $\ell_{1,\infty}$ has the lower $\ell_{2,2}$  error. However, $\ell_{2,2}$  error is not more relevant
for our purpose than any other norm.
   
\end{remark}

\begin{figure}[t]
    \centering    
    \includegraphics[width=0.49\textwidth,height=4.5cm]{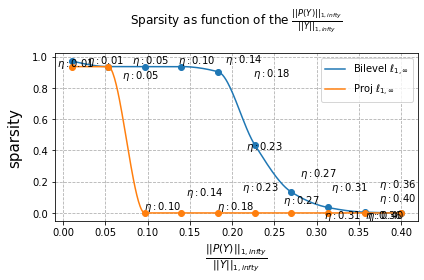} 
    \includegraphics[width=0.49\textwidth,height=4.5cm]{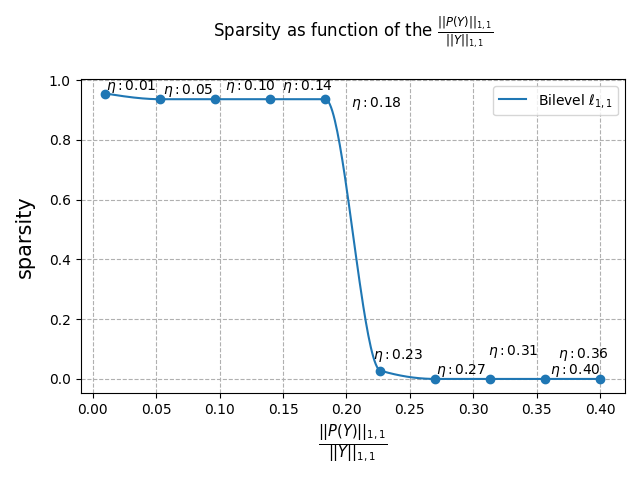}
    \includegraphics[width=0.49\textwidth,height=4.5cm]{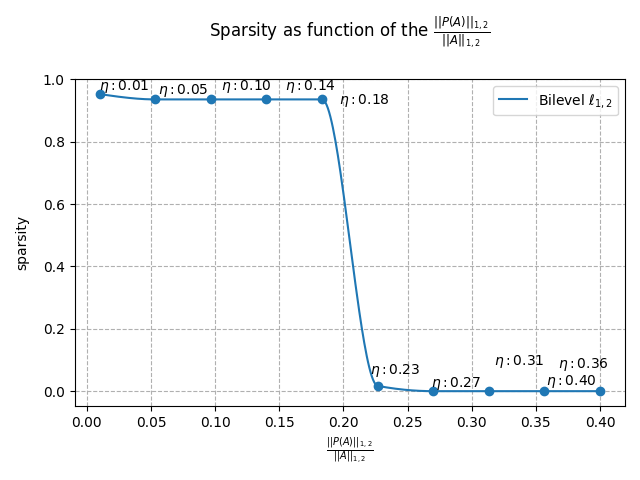}    
    \caption{64 informative features  Sparsity Top: the Bilevel $\ell_{1,\infty}$, Middle: Bilevel $\ell_{1,1}$, bottom: Bilevel $\ell_{1,2}$ projection}
    \label{sparsity}
\end{figure}

\begin{figure}[t]
    \centering    
    \includegraphics[width=0.49\textwidth,height=4.5cm]{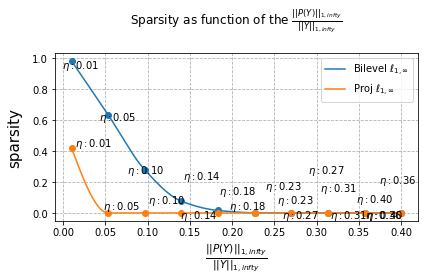} 
    \includegraphics[width=0.49\textwidth,height=4.5cm]{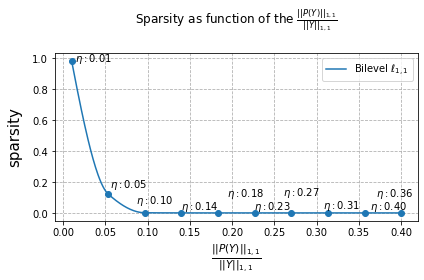}
    \includegraphics[width=0.49\textwidth,height=4.5cm]{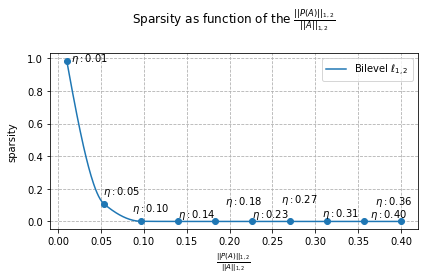}    
    \caption{16  informative features. Sparsity: Top the Bilevel $\ell_{1,\infty}$, Middle: Bilevel $\ell_{1,1}$, bottom: Bilevel $\ell_{1,2}$ projection}
    \label{sparsity-16}
\end{figure}

\begin{table}[ht]
    \centering  
    \begin{tabular}{|c|c|c|c|c|c|}
    \hline
      Cum-Sparsity ($\%$) & bilevel $\ell_{1,\infty}$ &bilevel $\ell_{1,1}$ &bilevel $\ell_{1,2}$ & $\ell_{1,\infty}$ \\
       
        \hline   
         datas-64&5.36  & 4.714&   4.705& 1.872\\
         \hline   
         data-16 &1.99 & 1.09&   1.07& 0.419\\
        \hline        
    \end{tabular}
    \caption{Comparison of Sparsity for two datasets with different informative features, where Cum-Sparsity ($\%$)) is the sum of sparsity over the test sets}.
    \label{Cumsparsity}
\end{table}

Table \ref{Cumsparsity} shows that our bilevel $\ell_{1,\infty}$ projection outperforms sparsity of the usual $\ell_{1,\infty}$ projection. Although the sparsity curves look very close. Table \ref{Cumsparsity} shows  that bilevel $\ell_{1,1}$ projection is slightly more sparse than bilevel $\ell_{1,2}$.
\clearpage
\newpage

The code is available online \footnote{https://github.com/MichelBarlaud/SAE-Supervised-Autoencoder-Omics}

\subsection{Experimental results on classification and  feature selection using a supervised autoencoder neural network}
\subsubsection{Supervised Autoencoder (SAE) framework}
Autoencoders were introduced within the field of neural networks decades ago, their most efficient application  being dimensionality reduction \cite{deep}. 
Autoencoders were used in application ranging from unsupervised deep-clustering \cite{VAE,Rochelle} to supervised learning, adding a classification loss in order to improve classification performance \cite{LESAE,ICASSP}.
In this paper, we use a supervised autoencoder with the cross entropy as the added classification loss.\\

Let $X$ be the concatenated raw data matrix ($n \times m$) (n is the number of samples (cells) and m the number of genes). 
Let $\widehat{X}$  the reconstructed data and $W$ the weights of the neural network.
Let $Z$ the encoded latent space. Note that the dimension of the latent space $k$ corresponds to the number of classes.\\
The goal is to learn the network weights, $W$ minimizing the total loss.
In order to sparsify the neural network,
we propose to use the different bi-level projection methods as a constraint to enforce sparsity in our model.
 The global criterion is:
 \begin{equation}
 \label{crit}
\underset{W}{\text{minimize}}  \quad \phi(X,Y) \quad \text{ subject to }  \quad BP^{1,\infty}(W) \leq \eta
\end{equation}
where $\phi(X,Y)=$ $\alpha  \cdot \psi (X,\widehat{X}) +$ $\mathcal{H}(Y,Z)$. 
We use the robust Smooth $\ell_1$ (Huber) Loss \cite{Huber} as the reconstruction loss $\psi$.
Parameter $\alpha$ is a linear combination factor used to define the final loss.  
We compute the mask by using the various bilevel projection methods, and we use the double descent algorithm \cite{Lottery,double} for minimizing the criterion \ref{crit}.
We implemented our SAE method using the PyTorch framework for the model, optimizer, schedulers and loss functions. We use a fully connected neural network with only one hidden layer (dimension 100) and a latent layer of dimension $k=2$. 
We chose the ADAM optimizer \cite{Adam}, as the standard optimizer in PyTorch. We use the smooth SiLU activation function.

\subsubsection{Experimental accuracy results on autoencoder neural networks }

\begin{figure}[ht]
    \centering
    \includegraphics[width=0.49\textwidth,height=4.cm]{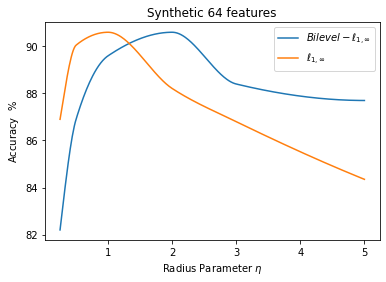}
    \includegraphics[width=0.49\textwidth,height=4.cm]{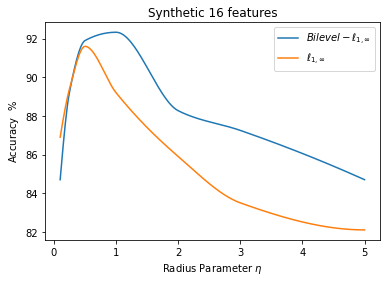}
    \caption{Accuracy  as a function of the radius parameter $\eta$; Top, 64 informative features :Bottom 16 informative features }
    \label{SyntRadius}
\end{figure}
Figure~\ref{SyntRadius} shows the impact of the radius ($\eta$) on a synthetic  dataset using the bilevel projection versus the usual projection.
It can be seen that the best accuracy is obtained for $\eta=0.5 $ for $\ell_{1,\infty}$ and for $\eta= 1$ for the bilevel $\ell_{1,\infty}$ projection. Moreover, the accuracy is more robust to parameter $\eta$ for the bilevel projection.
Table \ref{SyntAccuracysparsity} shows accuracy classification for 64 informative features. The baseline is an implementation that does not process any projection.
Compared to the baseline, the SAE using the $\ell_{1,\infty}$ projection improves the accuracy by $10.3 \%$.\\
Table \ref{SyntAccuracysparsity-16} shows that accuracy results for 16 informative features of the bilevel  $\ell_{1,\infty}$ and classical $\ell_{1,\infty}$ is slightly better.

From Tables \ref{SyntAccuracysparsity} on synthetic dataset it can be seen that the best accuracy is obtained for $\eta=0.5 $ for $\ell_{1,\infty}$ and for $\eta= 1. $ for the bilevel $\ell_{1,\infty}$ projection. Maximum accuracy of both method are similar. 
However, sparsity and computation time are better for bilevel $\ell_{1,\infty}$ than for regular $\ell_{1,\infty}$.

\begin{table}[ht]
    \centering  
    \begin{tabular}{|c|c|c|c|c|}
    \hline
      Synthetic 64 & Baseline &$\ell_{1,\infty}$ & bilevel $\ell_{1,\infty}$   \\
      \hline  
      Best Radius & -& 1 & 2.0 \\
      \hline   
        Accuracy  $ \%$ &80.3 $\pm 1.8$ &90.6.6$\pm 2.85$ & 90.6 $\pm 1.24 $ \\
        \hline   
        \hline       
    \end{tabular}
   \caption{\textbf{Synthetic } dataset 64 features.  SiLU activation, Accuracy : comparison of  $\ell_{1,\infty}$ and  bi-level $\ell_{1,\infty}$.}
   \label{SyntAccuracysparsity}
\end{table}

\begin{table}[ht]
    \centering  
    \begin{tabular}{|c|c|c|c|c|}
    \hline
      Synthetic 16 & Baseline &$\ell_{1,\infty}$ & bilevel $\ell_{1,\infty}$   \\
      \hline  
      Best Radius & -& 0.5 & 1.0 \\
      \hline   
        Accuracy  $ \%$ &74.6 $\pm 2.2$ &91.6 $\pm 3.3$& 92.36 $\pm 1.9 $ \\
        \hline   
      
    \end{tabular}
   \caption{\textbf{Synthetic } dataset 16 features.  SiLU activation, Accuracy : comparison of  $\ell_{1,\infty}$ and  bi-level $\ell_{1,\infty}$.}
   \label{SyntAccuracysparsity-16}
\end{table}

We consider the HIF2 dataset, now one of the real dataset from the study of in single-cell CRISPRi screening \cite{Crispri}. This HIF2 dataset is composed of 779 cells and 10,000 features (genes) \cite{Crispri}.  
\begin{figure}[ht]
    \centering
    \includegraphics[width=0.49\textwidth,height=4.cm]{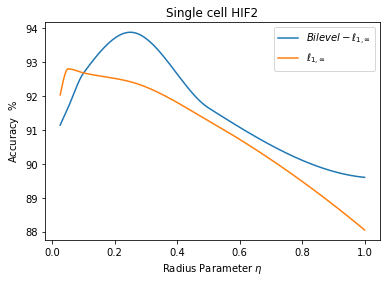}    
    \caption{Accuracy  as a function of the radius parameter $\eta$; HIF2 dataset}
    \label{HIF2Radius}
\end{figure}
Figure~\ref{HIF2Radius} shows the curve of the accuracy versus the radius ($\eta$) on HIF2 dataset. We can see that the curve is more well shaped using by the bilevel projection.

\begin{table}[ht]
    \centering  
    \begin{tabular}{|c|c|c|c|c|}
    \hline
      Real data HIF2 & Baseline &$\ell_{1,\infty}$ & bilevel $\ell_{1,\infty}$   \\
      \hline  
      Best Radius & -& 0.1 & 0.25 \\
      \hline   
        Accuracy  $ \%$ &84.6 $\pm 1.2$ &92.68 $\pm 1.85$& 93.88 $\pm 1.8 $ \\
        \hline   
      
    \end{tabular}
   \caption{\textbf{HIF2 dataset }   SILU activation, Accuracy : comparison of  $\ell_{1,\infty}$ and  bilevel $\ell_{1,\infty}$.}
   \label{Accuracysparsity-HIF2}
\end{table}
Again, table\ref{Accuracysparsity-HIF2} on this HIF2 real dataset, compared to the baseline the SAE using the bilevel $\ell_{1,\infty}$ projection improves the accuracy by $10 \%$ similarly to the results on the synthetic data-64.
Moreover, on this real dataset, our bilevel projection outperforms the usual $\ell_{1,\infty}$ projection by $1.2 \%$.

\begin{figure}[ht]
    \centering
    \includegraphics[width=0.49\textwidth,height=4.cm]{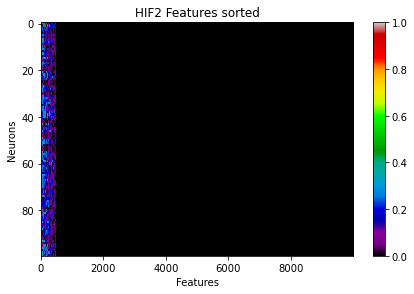}
    \caption{Weights of the first layer of the Fully connected Neural network}
    \label{sparsity}
\end{figure}

Constraint optimization  generally produces networks with random
sparse connectivity, i.e. sparse weight
matrices. They only reduce the memory cost but not the computational cost,
since they result in networks with sparse weight matrices.
Decreasing both
memory and computational requirements can however be
achieved by suppressing columns (features) instead of weights. Figure \ref{sparsity} shows that our bilevel $\ell_{1,\infty}$ suppress columns and thus efficiently reduces the computational cost.

\section{Discussion}
A first application of our bilevel $\ell_{1,\infty}$  projection is feature selection in biology \cite{Crispri}.\\ 
Second, our bilevel method can be extended straight forward to tensor and convolutionnal neural networks.
The $\ell_{1,\infty}$ projection has been successfully applied to the sparsification of autoencoders using convolutional neural networks with a memory reduction of 84 and a reduction in computational cost of 40 without visual image degradation \cite{CAE} in the new image compression standard JPEG  AI  \cite{JPEGAI}.\\
According to estimates, by 2040 Artificial Intelligence (AI) server farms may account for $14 ‰$ of all global carbon emissions in the world. 
Third, another application of our method can be the sparsification of attention matrices of a transformer architecture \cite{Transformer} used in AI software.

\section{Conclusion}

Although many projection algorithms
were proposed for the projection of the $\ell_{1,\infty}$  norm, complexity of these algorithms
remain an issue. The worst-case time complexity
of these algorithms is $\mathcal{O}\big(n \times m \times \log(n \times m)\big)$ for a matrix in $\RR^{n\times m}$. 
In order to cope with this complexity issue, we
have proposed a new bi-level  projection
method. The main motivation of our work is the
direct independent splitting made by the bi-level
optimization, which takes into account the structured
sparsity requirement. We showed that the
theoretical computational cost of our new bi-level
method is only $\mathcal{O}\big(n \times m \big)$ for a matrix in $\RR^{n\times m}$ and thus improves by $\mathcal{O}\big(\log(n m)\big)$ times classical algorithms. Note that this improvement can be huge for a large dataset.
Experiments on synthetic and a real data show that our
bi-level method is faster than the actual
fastest algorithm. Sparsity of  our bi-level $\ell_{1,\infty}$ projection  outperforms other bi-level projections $\ell_{1,1}$ and $\ell_{1,2}$.\\

\bibliography{references}

\begin{thebibliography}{10}
\providecommand{\url}[1]{#1}
\csname url@samestyle\endcsname
\providecommand{\newblock}{\relax}
\providecommand{\bibinfo}[2]{#2}
\providecommand{\BIBentrySTDinterwordspacing}{\spaceskip=0pt\relax}
\providecommand{\BIBentryALTinterwordstretchfactor}{4}
\providecommand{\BIBentryALTinterwordspacing}{\spaceskip=\fontdimen2\font plus
\BIBentryALTinterwordstretchfactor\fontdimen3\font minus \fontdimen4\font\relax}
\providecommand{\BIBforeignlanguage}[2]{{%
\expandafter\ifx\csname l@#1\endcsname\relax
\typeout{** WARNING: IEEEtran.bst: No hyphenation pattern has been}%
\typeout{** loaded for the language `#1'. Using the pattern for}%
\typeout{** the default language instead.}%
\else
\language=\csname l@#1\endcsname
\fi
#2}}
\providecommand{\BIBdecl}{\relax}
\BIBdecl

\bibitem{he2010stable}
Z.~He and W.~Yu, ``Stable feature selection for biomarker discovery,'' \emph{Computational biology and chemistry}, vol.~34, no.~4, pp. 215--225, 2010.

\bibitem{AIgreen}
R.~Schwartz, J.~Dodge, N.~A. Smith, and O.~Etzioni, ``{Green AI},'' 2019, preprint arXiv:1907.10597.

\bibitem{Alv2016}
J.~M. Alvarez and M.~Salzmann, ``Learning the number of neurons in deep networks,'' in \emph{Advances in Neural Information Processing Systems}, 2016, pp. 2270--2278.

\bibitem{Han2015-2}
S.~Han, J.~Pool, J.~Tran, and W.~Dally, ``Learning both weights and connections for efficient neural network,'' in \emph{Advances in neural information processing systems}, 2015, pp. 1135--1143.

\bibitem{Gom2019}
A.~N. Gomez, I.~Zhang, K.~Swersky, Y.~Gal, and G.~E. Hinton, ``Learning sparse networks using targeted dropout,'' \emph{arXiv :1905.13678}, 2019.

\bibitem{dropout1}
N.~Srivastava, G.~Hinton, A.~Krizhevsky, I.~Sutskever, and R.~Salakhutdinov, ``Dropout: a simple way to prevent neural networks from overfitting,'' \emph{The journal of machine learning research}, vol.~15, no.~1, pp. 1929--1958, 2014.

\bibitem{dropout2}
J.~Cavazza, P.~Morerio, B.~Haeffele, C.~Lane, V.~Murino, and R.~Vidal, ``Dropout as a low-rank regularizer for matrix factorization,'' in \emph{International Conference on Artificial Intelligence and Statistics (AISTATS)}, 2018, pp. 435--444.

\bibitem{tRS}
R.~Tibshirani, ``Regression shrinkage and selection via the lasso,'' \emph{Journal of the Royal Statistical Society. Series B (Methodological)}, pp. 267--288, 1996.

\bibitem{sparse}
T.~Hastie, R.~Tibshirani, and M.~Wainwright, ``Statistcal learning with sparsity: The lasso and generalizations,'' \emph{CRC Press}, 2015.

\bibitem{Yua}
M.~Yuan and Y.~Lin, ``Model selection and estimation in regression with grouped variables,'' \emph{Journal of the Royal Statistical Society: Series B (Statistical Methodology)}, vol.~68, no.~1, pp. 49--67, 2006.

\bibitem{Hua2018}
Z.~Huang and N.~Wang, ``Data-driven sparse structure selection for deep neural networks,'' in \emph{Proceedings of the European Conference on Computer Vision (ECCV)}, 2018, pp. 304--320.

\bibitem{Yoo2017}
J.~Yoon and S.~J. Hwang, ``Combined group and exclusive sparsity for deep neural networks,'' in \emph{Proceedings of the 34th International Conference on Machine Learning-Volume 70}.\hskip 1em plus 0.5em minus 0.4em\relax JMLR. org, 2017, pp. 3958--3966.

\bibitem{Sca2017}
S.~Scardapane, D.~Comminiello, A.~Hussain, and A.~Uncini, ``Group sparse regularization for deep neural networks,'' \emph{Neurocomputing}, vol. 241, pp. 81--89, 2017.

\bibitem{BCD}
N.~Simon, J.~Friedman, T.~Hastie, and R.~Tibshirani, ``A sparse-group lasso,'' \emph{Journal of Computational and Graphical Statistics}, vol.~22, no.~2, pp. 231--245, 2013.

\bibitem{Fastlasso}
I.~Yasutoshi, F.~Yasuhiro, and K.~Hisashi, ``Fast sparse group lasso,'' in \emph{Advances in Neural Information Processing Systems}, vol.~32.\hskip 1em plus 0.5em minus 0.4em\relax Curran Associates, Inc., 2019.

\bibitem{fht}
J.~Friedman, T.~Hastie, and R.~Tibshirani, ``Regularization path for generalized linear models via coordinate descent,'' \emph{Journal of Statistical Software}, vol.~33, pp. 1--122, 2010.

\bibitem{myCA}
J.~Mairal and B.~Yu, ``Complexity analysis of the lasso regularization path,'' in \emph{Proceedings of the 29th International Conference on Machine Learning (ICML-12)}, 2012, pp. 353--360.

\bibitem{Pesquet}
G.~Chierchia, N.~Pustelnik, J.~C. Pesquet, and B.~Pesquet-Popescu, ``Epigraphical projection and proximal tools for solving constrained convex optimization problems,'' \emph{Signal, Image and Video Processing}, 2015.

\bibitem{BBCF}
M.~Barlaud, W.~Belhajali, P.~Combettes, and L.~Fillatre, ``Classification and regression using an outer approximation projection-gradient method,'' \emph{IEEE TRANSACTIONS ON SIGNAL PROCESSING}, vol.~65, no.~17, pp. 4635--4643, 2017.

\bibitem{condat}
L.~Condat, ``Fast projection onto the simplex and the l1 ball,'' \emph{Mathematical Programming Series A}, vol. 158, no.~1, pp. 575--585, 2016.

\bibitem{Perez19}
G.~Perez, M.~Barlaud, L.~Fillatre, and J.-C. R{\'e}gin, ``A filtered bucket-clustering method for projection onto the simplex and the $\ell_1$-ball,'' \emph{Mathematical Programming}, May 2019.

\bibitem{quattoni2009efficient}
A.~Quattoni, X.~Carreras, M.~Collins, and T.~Darrell, ``An efficient projection for $\ell_{1,\infty}$ regularization,'' in \emph{Proceedings of the 26th International Conference on Machine Learning}, 2009, pp. 857--864.

\bibitem{bejar2021fastest}
B.~Bejar, I.~Dokmani{\'c}, and R.~Vidal, ``The fastest $\ell_{1,\infty}$ prox in the {W}est,'' \emph{IEEE transactions on pattern analysis and machine intelligence}, vol.~44, no.~7, pp. 3858--3869, 2021.

\bibitem{chau2019efficient}
G.~Chau, B.~Wohlberg, and P.~Rodriguez, ``Efficient projection onto the $\ell_{1,\infty}$ mixed-norm ball using a newton root search method,'' \emph{SIAM Journal on Imaging Sciences}, vol.~12, no.~1, pp. 604--623, 2019.

\bibitem{chu2020semismooth}
D.~Chu, C.~Zhang, S.~Sun, and Q.~Tao, ``Semismooth newton algorithm for efficient projections onto $\ell_{1,\infty}$-norm ball,'' in \emph{International Conference on Machine Learning}, 2020, pp. 1974--1983.

\bibitem{moreau62}
J.~J. Moreau, ``Fonctions convexes duales et points proximaux dans un espace hilbertien,'' \emph{Comptes Rendus de l'Acad{\'emie} des Sciences de Paris}, vol. A255, no.~22, pp. 2897--2899, Nov 1962.

\bibitem{bau17}
H.~H. Bauschke and P.~L. Combettes, \emph{Convex Analysis and Monotone Operator Theory in Hilbert Spaces}, 2nd~ed.\hskip 1em plus 0.5em minus 0.4em\relax New York: Springer, 2017.

\bibitem{Combettes}
P.~L. Combettes and J.-C. Pesquet, ``Proximal splitting methods in signal processing,'' in \emph{Fixed-point algorithms for inverse problems in science and engineering}.\hskip 1em plus 0.5em minus 0.4em\relax Springer, 2011, pp. 185--212.

\bibitem{con23}
L.~Condat, D.~Kitahara, A.~Contreras, and A.~Hirabayashi, ``Proximal splitting algorithms for convex optimization: A tour of recent advances, with new twists,'' \emph{SIAM Review}, vol.~65, no.~2, pp. 375--435, May 2023.

\bibitem{Perez-2023}
G.~Perez, L.~Condat, and M.~Barlaud, ``Near-linear time projection onto the l1,infty ball application to sparse autoencoders.'' \emph{IEEE International Conference on Tools with Artificial Intelligence Washington USA}, 2024.

\bibitem{bilevel2}
A.~Sinha, P.~Malo, and K.~Deb, ``A review on bilevel optimization: From classical to evolutionary approaches and applications,'' \emph{IEEE Transactions on Evolutionary Computation}, vol.~22, no.~2, pp. 276--295, 2018.

\bibitem{bilevel3}
K.~Bennett, J.~Hu, X.~Ji, G.~Kunapuli, and J.-S. Pang, ``Model selection via bilevel optimization,'' \emph{IEEE International Conference on Neural Networks - Conference Proceedings}, 2006.

\bibitem{Lcurves}
P.~C. Hansen and D.~P. O’Leary, ``The use of the l-curve in the regularization of discrete ill-posed problems,'' \emph{SIAM Journal on Scientific Computing}, vol.~14, 1993.

\bibitem{Moreau}
J.~Moreau, ``Proximit\'{e} et dualit\'{e} dans un espace hilbertien,'' \emph{Bull. Soc.Math. France., 93}, pp. 273--299, 1965.

\bibitem{Boyd}
N.~Parikh and S.~Boyd, ``Proximal algorithms,'' \emph{Foundations and Trends® in Optimization}, 2014.

\bibitem{deep}
I.~Goodfellow, Y.~Bengio, and A.~Courville, \emph{Deep learning}.\hskip 1em plus 0.5em minus 0.4em\relax MIT press, 2016, vol.~1.

\bibitem{VAE}
D.~Kingma and M.~Welling, ``Auto-encoding variational bayes,'' \emph{International Conference on Learning Representation}, 2014.

\bibitem{Rochelle}
J.~Snoek, R.~Adams, and H.~Larochelle, ``On nonparametric guidance for learning autoencoder representations,'' in \emph{Artificial Intelligence and Statistics}.\hskip 1em plus 0.5em minus 0.4em\relax PMLR, 2012, pp. 1073--1080.

\bibitem{LESAE}
L.~Le, A.~Patterson, and M.~White, ``Supervised autoencoders: Improving generalization performance with unsupervised regularizers,'' \emph{Advances in Neural Information Processing Systems}, 2018.

\bibitem{ICASSP}
M.~Barlaud and F.~Guyard, ``Learning a sparse generative non-parametric supervised autoencoder,'' \emph{Proceedings of the International Conference on Acoustics, Speech and Signal Processing, Toronto, Canada}, June 2021.

\bibitem{Huber}
P.~J. Huber, \emph{Robust statistics}.\hskip 1em plus 0.5em minus 0.4em\relax Wiley, New York, 1981.

\bibitem{Lottery}
J.~Frankle and M.~Carbin, ``The lottery ticket hypothesis: Finding sparse, trainable neural networks,'' \emph{arXiv preprint arXiv:1803.03635}, 2018.

\bibitem{double}
H.~Zhou, J.~Lan, R.~Liu, and J.~Yosinski, ``Deconstructing lottery tickets: Zeros, signs, and the supermask,'' in \emph{Advances in Neural Information Processing Systems 32}, 2019, pp. 3597--3607.

\bibitem{Adam}
D.~Kingma and J.~Ba, ``a method for stochastic optimization.'' \emph{International Conference on Learning Representations}, pp. 1--13, 2015.

\bibitem{Crispri}
M.~Truchi, C.~Lacoux, C.~Gille, J.~Fassy, V.~Magnone, R.~Lopes~Goncalves, C.~Girard-Riboulleau, I.~Manosalva-Pena, M.~Gautier-Isola, K.~Lebrigand, P.~Barbry, S.~Spicuglia, G.~Vassaux, R.~Rezzonico, M.~Barlaud, and B.~Mari, ``Detecting subtle transcriptomic perturbations induced by lncrnas knock-down in single-cell crispri screening using a new sparse supervised autoencoder neural network,'' \emph{Frontiers in Bioinformatics}, 2024.

\bibitem{CAE}
G.~Cyprien, F.~Guyard, M.~Antonini, and M.~Barlaud, ``Learning sparse autoencoders for green ai image coding,'' \emph{Proceedings of the International Conference on Acoustics, Speech and Signal Processing, Rhodes, Greece}, June 2023.

\bibitem{JPEGAI}
J.~Ascenso, E.~Alshina, and T.~Ebrahimi, ``The jpeg ai standard: Providing efficient human and machine visual data consumption,'' \emph{IEEE MultiMedia}, vol.~30, no.~1, pp. 100--111, 2023.

\bibitem{Transformer}
A.~Vaswani, N.~Shazeer, N.~Parmar, J.~Uszkoreit, L.~Jones, A.~N. Gomez, L.~Kaiser, and I.~Polosukhin, ``Attention is all you need,'' \emph{Advances in Neural Information Processing Systems}, 2017.

\end{thebibliography}
\bibliographystyle{IEEEtran}

\end{document}